\documentclass{bmvc2k}

\usepackage{booktabs}
\usepackage{multirow}
\usepackage[noend]{algpseudocode}
\usepackage{color}
\usepackage[linesnumbered,ruled,vlined]{algorithm2e}
\usepackage{amsmath}
\usepackage{amssymb}
\usepackage{dsfont}
\usepackage{wrapfig,lipsum,booktabs}

\SetKwInput{KwInput}{Input}                % Set the Input
\SetKwInput{KwOutput}{Output}              % set the Output

\definecolor{pink}{rgb}{0.858, 0.188, 0.478}

\DeclareMathOperator*{\argmax}{arg\,max}

\title{PropMix: Hard Sample Filtering and Proportional MixUp for Learning with Noisy Labels}

\addauthor{Filipe R. Cordeiro}{filipe.rolim@ufrpe.br}{1}
\addauthor{Vasileios Belagiannis}{vasileios.belagiannis@uni-ulm.de}{2}
\addauthor{Ian Reid}{ian.reid@adelaide.edu.au}{3}
\addauthor{Gustavo Carneiro}{gustavo.carneiro@adelaide.edu.au}{3}

\addinstitution{
 Universidade Federal Rural \\
 de Pernambuco\\
 Recife, Brazil
}
\addinstitution{
 Universität Ulm\\ 
 Ulm, Germany \\
}
\addinstitution{
 University of Adelaide\\
 Adelaide, Australia
}

\runninghead{CORDEIRO ET AL.}{PropMix: Hard Sample Filtering and Proportional MixUp}

\begin{document}

\maketitle

\begin{abstract}
The most competitive noisy label learning methods rely on an unsupervised classification of clean and noisy samples, where
samples classified as noisy are re-labelled and "MixMatched" with the clean samples.
These methods have two issues in large noise rate problems: 1) the noisy set is more likely to contain hard samples that are incorrectly re-labelled, and 2) the number of samples produced by MixMatch tends to be reduced because it is constrained by the small clean set size.
In this paper, we introduce the learning algorithm PropMix to handle the issues above. PropMix filters out hard noisy samples, with the goal of increasing the likelihood of correctly re-labelling the easy noisy samples.
Also, PropMix places clean and re-labelled easy noisy samples in a training set that is augmented with MixUp, removing the clean set size constraint and including a large proportion of correctly re-labelled easy noisy samples. 
We also include self-supervised pre-training to improve robustness to high noisy label scenarios.
Our experiments show that PropMix has state-of-the-art (SOTA) results on CIFAR-10/-100 (with symmetric, asymmetric and semantic 
label noise), Red Mini-ImageNet (from the Controlled Noisy Web Labels), Clothing1M and  WebVision. 
In severe label noise benchmarks, our results are substantially better than other methods. The code is available at \url{https://github.com/filipe-research/PropMix}.
\end{abstract}

%-------------------------------------------------------------------------
\section{Introduction}
\label{sec:intro}

Deep neural network models reached promising results in several computer vision applications recently~\citep{aggarwal2021diagnostic, minaee2021image, grigorescu2020survey}. Nevertheless, the superior performance depends on the availability of well-curated large-scale training sets with clean annotations~\citep{litjens2017survey}. The labeling process can produce noisy labels due to human failure, low-quality data, and challenging labelling tasks~\citep{frenay_survey}. The problem is that noisy labels in the training set harms the learning process by reducing model generalization~\citep{zhang2016understanding}. Developing methods that are robust to label noise is important to deal with real-world applications, where noisy annotations is often part of the training set.

A common approach to address this challenge is based on semi-supervised learning (SSL) methods~\citep{ding2018semi,ortego2019towards,ortego2020multi,DivideMix}, based on a 2-stage process formed by an unsupervised learning method to classify training samples as clean or noisy, followed by SSL that "MixMatches"~\citep{MixMatch} the labelled set formed by the samples classified as clean, and the re-labelled samples from the noisy set.
There are two issues with such approach.  First, samples classified as noisy can be easy or hard to be re-labelled, where the hard samples are unlikely to be correctly re-labelled, which can bias the training process.
Second, MixMatch~\citep{MixMatch} relies on a one-to-one sampling between the clean and noisy sets, where the total number of samples is constrained by the clean set size. However, the clean set tends to be smaller and the noisy set tends to have more incorrectly re-labelled samples for larger noise rates--such fact impairs the robustness of SSL methods for large noise rate problems.
We hypothesise that by filtering out hard noisy samples, we can reduce the risk of over-fitting those samples.
Furthermore, by re-labelling the easy noisy samples and mixing them with the clean set with MixUp~\citep{mixup} for training a classifier, we not only remove the constraint on the clean set size, but also use a noisy set with better chances to have correctly re-labelled samples, allowing the method to be more robust to large noise rate problems. 

In this paper, we propose a new learning algorithm called PropMix that addresses the two points above. 
PropMix filters out hard noisy samples via a two-stage process, where the first stage classifies samples as clean or noisy using the loss values, and the second stage eliminates hard noisy samples using their classification confidence. 
Then, by re-labelling the easy noisy samples with the model output, adding these samples to the training set, and running a regular classification training with MixUp~\citep{mixup}, we  show that PropMix is robust to a wide range of noise rates.
To improve the feature representation and model confidence in high noise scenarios, we also add a self-supervised pre-training stage~\citep{MoCo,MoCoV2,SCAN,SimCLR}.
Empirical results on CIFAR-10/-100~\citep{krizhevsky2009learning} under symmetric, asymmetric and semantic 
noise, show that PropMix outperforms previous approaches. For high-noise rate problems in CIFAR-10/-100~\citep{DivideMix} and 
Red Mini-ImageNet from the Controlled Noisy Web Labels~\citep{FaMUS}, PropMix presents the best results in the field by a substantial margin.

\vspace{-.1in}
\section{Prior Work}

Recently proposed noisy label learning methods rely on the following strategies: \textit{robust loss} functions~\citep{ma2020normalized,wang2019imae, wang2019symmetric}, \textit{sample selection}~\citep{han2018co}, \textit{label cleansing} \cite{jaehwan2019photometric, yuan2018iterative}, \textit{sample weighting} \cite{ren2018learning}, \textit{meta-learning}~\citep{han2018pumpout}, \textit{ensemble learning} \cite{miao2015rboost} and \textit{semi-supervised learning} (SSL)~\citep{MixMatch}. 
The most successful approaches use SSL, combined with other methods~\citep{jiang2018mentornet,DivideMix,elr2020}.

SSL methods~\citep{DivideMix,ortego2019towards} first classify samples as clean or noisy, where the noisy samples are re-labelled by the model, and these clean and noisy sets are combined with MixMatch~\citep{MixMatch}.
As mentioned before, SSL methods have two issues:
1) the training set size for the MixMatch stage is limited by the clean set size that reduces with increasing label noise, and 2) the noisy sample re-labelling accuracy also reduces with increasing label noise. 
Although the first point is not addressed by SSL methods, the second point is mitigated by replacing the cross entropy (CE) loss by the Mean Absolute Error (MAE) loss to fit the noisy samples.
Although MAE has been shown to be robust to label noise~\citep{ghosh2017robust}, it tends to underfit the training set~\citep{ma2020normalized}. 

An alternative way to reduce the risk of overfitting incorrectly re-labelled noisy samples is by rejecting them altogether~\citep{han2018co,pleiss2020identifying}, but such rejection ignores the information present in noisy samples, which can decrease the effectiveness of the approach. Moreover, in high noise rate scenarios, the filtered clean set can be too small to train the model. 
The noisy set can be used after re-labelling the noisy samples~\citep{proselflc,DivideMix,ortego2019towards}, which works well for low noise rate problems, but as the noise rate increases, this approach is not effective given that the incorrectly re-labelled noisy samples tend to bias the training.
The use of clean validation sets in a meta-learning approach~\citep{algan2021meta} can reduce this issue, but the existence of a clean validation set may be infeasible in real world applications.
SELFIE~\citep{selfie} proposes label correction to a subset of samples that present consistent prediction, while discarding samples that are less consistent. The main issue with SELFIE is that the classification of clean samples is based on a loss threshold that might include some relabeling in the predicted clean set, 
% \gustavo{I don't get this point... do they re-label samples classified as clean?} \filipe{[According to the paper "...not necessarily true that
% $R \cap C =\varnothing $. If a sample $x \in R \cap C$ , being refurbishable
% precedes being clean because mislabeled instances could
% be included even in $C$", where C is the clean set and R the refurbishable/relabelled set ]}
producing false positives. Our method proposes a hybrid approach. We claim that hard noisy samples are unlikely to have their label corrected, mainly in a high noise scenario. On the other hand, we can find easy noisy samples that are likely to be correctly relabelled and used in a supervised training. The main difference of existing filtering methods and our approach is that we filter out hard noisy samples, while keeping easy noisy samples to be relabelled and included in the training process. We show in the experiments that easy noisy samples can be relabelled correctly with high accuracy, whereas hard noisy samples are unlikely to be correctly relabelled and therefore should be removed from training.

\vspace{-.1in}

\section{Method}

\subsection{Problem Definition}
\label{sec:problem_definition}

Consider the training set be denoted by $\mathcal{D}=\{(\mathbf{x}_i, \mathbf{y}_i)\}_{i=1}^{|\mathcal{D}|}$, with  $\mathbf{x}_i \in \mathcal{S} \subset \mathbb{R}^{H \times W \times 3}$ being the $i^{th}$ RGB image of size $H \times W$, and $\mathbf{y}_i \in \{0,1\}^{|\mathcal{Y}|}$ denoting a one-hot vector representing the given label, with $\mathcal{Y} \in \{1,...,|\mathcal{Y}|\}$ denoting the set of labels, and $\sum_{c \in \mathcal{Y}} \mathbf{y}_i(c)=1$. The hidden true label $\hat{\mathbf{y}}_i$ can differ from the given label  $\mathbf{y}_i$ that is a result of noise process, represented by $\mathbf{y}_i \sim p(\mathbf{y} | \mathbf{x}_i,\mathcal{Y},\hat{\mathbf{y}}_i)$, with $p(\mathbf{y}(j) | \mathbf{x}_i,\mathcal{Y},\hat{\mathbf{y}}_i(c))=\eta_{jc}(\mathbf{x}_i)$,
where the $j,c\in\mathcal{Y}$ are the classes, $\eta_{jc}(\mathbf{x}_i) \in [0,1]$ the probability of flipping the class $c$ to $j$, and $\sum_{j \in \mathcal{Y}}\eta_{jc}(\mathbf{x}_i)=1$. There are three common types of noise in the literature:  %We assume that this noise process can be of three types, namely
symmetric~\citep{kim2019nlnl}, asymmetric~\citep{patrini2017making}, and semantic~\citep{rog}.
The symmetric noise is a noisy type where the hidden label are flipped to a 
% The symmetric noise, also called uniform noise, refers to a noise type that the hidden label flips to a 
random class
with a fixed probability $\eta$, where the true label is included into the label flipping options, which means that $\eta_{jc}(\mathbf{x}_i)=\frac{\eta}{|\mathcal{Y}|-1}, \forall j \in \mathcal{Y}, \text{ such that } j \neq c$, and $\eta_{cc}(\mathbf{x}_i)=1-\eta$. 
The asymmetric noise has its labels flipped %is based on flipping labels 
between similar-looking object categories~\citep{patrini2017making}, where $\eta_{jc}(\mathbf{x}_i)$ depends only on the classes $j,c\in\mathcal{Y}$, but not on $\mathbf{x}_i$. %For example, using CIFAR-10 data set \cite{krizhevsky2009learning}, the asymmetric noise maps  \emph{bird} $\to$ \emph{plane}, \emph{truck} $\to$ \emph{automobile}, \emph{deer} $\to$ \emph{horse}, as in \cite{zhang2018generalized}. 
Finally, the semantic noise~\citep{rog} is the noisy type where the label flipping depends both on the classes $j,c\in\mathcal{Y}$ and image $\mathbf{x}_i$. % with similar semantic features. 

\vspace{-.1in}
\subsection{PropMix}
\label{sec:propmix}

\begin{figure}[!t]
\centering
\includegraphics[width=0.8\columnwidth]{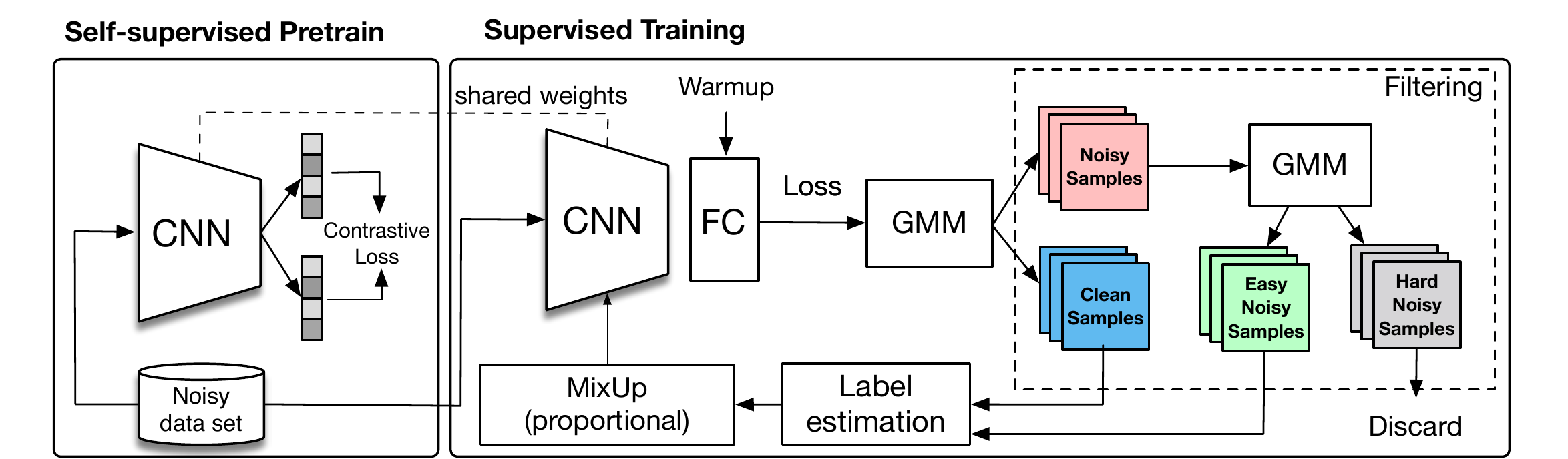}
% \vspace{0.1in}
\caption{Our proposed PropMix has a self-supervised pre-training stage~\citep{SimCLR,MoCo,MoCoV2,SCAN}, followed by a supervised training stage, where we first warm-up the classifier with a classification loss, using the pre-trained weights. Then, using the classification loss, we train a GMM to separate the samples into clean and noisy. Next, using classification confidence for the noisy set, we train a second GMM to separate the easy and hard noisy samples. The clean and easy noisy samples are proportionally combined in the MixUp for training.  }
\label{fig:propmix}
\end{figure}

The proposed PropMix algorithm (Fig.~\ref{fig:propmix}) starts with a self-supervised pre-training ~\citep{SimCLR,MoCo,MoCoV2,SCAN}.  Then, we perform a supervised training, with a new filtering step to identify clean samples, easy noisy samples, and hard noisy samples, which are removed from training. The easy noisy samples are re-labelled and proportionally combined with clean samples using MixUp~\citep{mixup} for supervised training. 

The self-supervised pre-training estimates $\phi$ of the feature representation $f_{\phi}:\mathcal{S} \to \mathcal{Z} \subset \mathbb{R}^d$, by minimizing the contrastive loss~\citep{SimCLR,MoCo,MoCoV2}:
\begin{equation}
    % \min_{\phi}{d(f_{\phi}(\mathbf{x}), f_{\phi}(\mathbf{x'}))}.
    \ell_{i,j}= -log\frac{exp(sim(\mathbf{z}_i, \mathbf{z}_j)/\tau)}{\sum_{k=1}^{2N} \mathds{1}_{[k\neq i]} exp(sim(\mathbf{z}_i, \mathbf{z}_k)/\tau) }
\end{equation}
with $N$ denoting mini-batch size, 
$\mathbf{z}=f_{\phi}(\mathbf{x})$
represents the features extracted from input $\mathbf{x}$, with $\mathbf{z_i}$ and  $\mathbf{z_j}$ being the feature vectors from two views of the same image (these views are obtained via different data augmentations of the same image),
$\mathds{1}_{[k\neq i]}\in\{0,1\}$ being %\sout{a one-hot vector}
an indicator function, $\tau$ denoting the temperature parameter, and $sim(.)$ representing the cosine similarity.
Following ~\citep{SCAN}, we then learn a clustering classifier $p_{\theta}(.|f_{\phi}(\mathbf{x}))$. More precisely, we form an initial set of $K$ nearest neighbours (KNN) in $\mathcal{Z}$ for each training sample, producing the set $\mathcal{N}_{\mathbf{x}_{i}} = \{ \mathbf{x}_j \}_{j=1}^K$ (for $\mathbf{x}_j \in \mathcal{D}$) for each sample $\mathbf{x}_i \in \mathcal{D}$ . We train $p_{\theta}(.|f_{\phi}(\mathbf{x}))$ with~\citep{SCAN}:
\begin{equation}
    \ell_{CLU}= \ell_{\mathcal{N}} +
    \lambda_e\ell_{e},
    \label{eq:cluster_loss}
\end{equation}
with 
$\ell_{\mathcal{N}} = -\frac{1}{|\mathcal{D}|} \sum_{i=1}^{|\mathcal{D}|} \sum_{\mathbf{x}_j \in \mathcal{N}_{\mathbf{x}_{i}}} q(z_{ji}) \log \left ( p_{\theta}(.|f_{\phi}(\mathbf{x}_i))^{\top}p_{\theta}(.|f_{\phi}(\mathbf{x}_j)) \right )$,
$q(z_{ji}) = 1$ if $\mathbf{x}_i$ and $\mathbf{x}_j$ have the same classification result (i.e.,
$\arg\max_{c \in \mathcal{Y}}p_{\theta}(c | f_{\phi}(\mathbf{x}_i))=\arg\max_{c \in \mathcal{Y}}p_{\theta}(c | f_{\phi}(\mathbf{x}_j))$),
and 
$\ell_e=\frac{1}{|\mathcal{D}|}
\sum_{c \in \mathcal{Y}}\sum_{\mathbf{x} \in \mathcal{D}}p_{\theta}(c|f_{\phi}(\mathbf{x})) \log p_{\theta}(c|f_{\phi}(\mathbf{x}))$
that maximises the entropy of the average classification and is weighted by $\lambda_e$. 
The weights from the feature map $f_{\phi}(\mathbf{x})$ are also updated in the clustering process using backpropagation.

After the pre-training,  we warm-up the classifier by training it for a few epochs on the (noisy) training data set with the cross-entropy (CE) loss. The clean and noisy sets, $\mathcal{X},\mathcal{U} \subseteq \mathcal{D}$, are formed with~\citep{arazo2019unsupervised, DivideMix,rog, jiang2020beyond}:
\begin{equation}
%\begin{split}
    \mathcal{X}  = \left \{ (\mathbf{x}_i,\mathbf{y}_i, w_i) : w_i=p \left ( \text{clean} | \ell_i , \gamma \right ) \ge \tau \right \}, \quad
    \mathcal{U}  = \left \{ (\mathbf{x}_i,\mathbf{y}_i, w_i) : w_i=
p \left ( \text{clean} | \ell_i , \gamma \right ) < \tau \right \},
%\end{split}
\label{eq:clean_noisy_sets}
\end{equation}
with $\tau$ denoting a classification threshold,
$\ell_i = -\mathbf{y}_i^{\top}\log p_{\theta}(.|f_{\phi}(\mathbf{x}_i))$, and $p \left ( \text{clean} | \ell_i , \gamma \right )$ being a function that estimates the probability that $(\mathbf{x}_i,\mathbf{y}_i)$ is a clean label sample.  The function $p \left ( \text{clean} | \ell_i , \gamma \right )$ in Eq.\ref{eq:clean_noisy_sets} is a bi-modal Gaussian mixture model (GMM)~\citep{DivideMix}, where $\gamma$ denotes the GMM parameters and the larger mean component is the noisy component whereas the smaller mean component is the clean component. Next, we obtain the sets of easy and hard noisy samples $\mathcal{U}_{E},\mathcal{U}_H \subseteq \mathcal{U}$, as follows: 
%\\
%. The mutually exclusive subsets $\mathcal{U}_{E}, \mathcal{U}_H \subseteq \mathcal{U}$ are formed with:
%\scalebox{0.85}{
\begin{equation}
\begin{split}
    \mathcal{U}_{E}  &= \left \{ (\mathbf{x}_i,\mathbf{y}_i) : p \left ( \text{hard} |  p_{\theta}(c_i^{*}|f_{\phi}(\mathbf{x})_i),  \gamma \right ) < \tau' \right \}, \\
    \mathcal{U}_{H}  &= \left \{ (\mathbf{x}_i,\mathbf{y}_i) : 
p \left ( \text{hard} | p_{\theta}(c_i^{*}|f_{\phi}(\mathbf{x}_i)) , \gamma \right ) \ge \tau' \right \},
\end{split}
\label{eq:hard_noisy_set}
\end{equation}
%}\\
%\\
with $c_i^{*}=\argmax_{c_i \in \mathcal{Y}} p_{\theta}(c_i|f_{\phi}(\mathbf{x}_i ))$, $p \left ( \text{hard} | p_{\theta}(c_i^{*}|f_{\phi}(\mathbf{x}_i)) , \gamma \right )$ being a function that estimates the probability that $(\mathbf{x}_i,\mathbf{y}_i)$ is a hard noisy label sample, and $\tau'$ denoting the hard noisy sample threshold.  The function $p \left ( \text{hard} |  p_{\theta}(c_i^{*}|f_{\phi}(\mathbf{x}_i)),\gamma \right)$ in Eq.~\ref{eq:hard_noisy_set} is a GMM, where $\gamma$ denotes the GMM parameters and the smaller mean component is the hard noise component whereas the larger mean component is the easy noise component. We assume hard noisy samples have wrong label and wrong prediction, so we remove them from training.

Next, we perform $M$ data augmentations based on geometrical and visual transformations to increase the number of samples in $\mathcal{X}$ and $\mathcal{U}_{E}$, generating the augmented sets $\mathcal{X}'=\{(\hat{\mathbf{x}}_i,\hat{\mathbf{y}}_i)\}_{i=1}^{|\mathcal{X}'|}$ and $\mathcal{U}_E'=\{(\hat{\mathbf{u}}_i,\hat{\mathbf{q}}_i)\}_{i=1}^{|\mathcal{U}_E'|}$, with $\hat{\mathbf{x}}_i$ and $\hat{\mathbf{u}}_i$ being the augmented samples from $\mathcal{X}$ and $\mathcal{U}_E$, respectively, and $\hat{\mathbf{y}}_i$ and $\hat{\mathbf{q}}_i$ the label estimations defined as follows:
\begin{equation}
    \hat{\mathbf{y}}_i = \text{TempShrp}(w_i \mathbf{y}_i + (1-w_i)\mathbf{p}_b;T),  \quad 
                \hat{\mathbf{q}}_b = \text{TempShrp}(\mathbf{q}_b;T)
\end{equation}
with  $\mathbf{p}_b =  \frac{1}{M}\sum\limits_{m=1}^{M}
p_{\theta}(.|f_{\phi}(\hat{\mathbf{x}}_{i,m}))$,  
$\mathbf{q}_b = \frac{1}{M}\sum\limits_{m=1}^{M} p_{\theta}(.|f_{\phi}(\hat{\mathbf{u}}_{i}))$ , and TempShrp(.) being a temperature sharpening~\citep{DivideMix}.

The linear combination between the clean and easy noisy samples rely on MixUp~\citep{mixup} data augmentation, which is different from the
SOTA SSL approaches~\citep{DivideMix} that use MixMatch~\citep{MixMatch} which has a training set size restricted by the clean set size. 
In particular, MixMatch-based methods select $B_{\mathcal{U}'}=|\mathcal{X}'|$ samples from the noisy set and $B_{\mathcal{X}'} = |\mathcal{X}'|$ samples from the clean set, so the MixUp proportion of samples from the noisy set is $\zeta=\frac{B_{\mathcal{U}'}}{{B_{\mathcal{X}'}+B_{\mathcal{U}'}}}=0.5$, and the size of the MixMatch training set is $|\mathcal{X}'|$. 
In our approach, we re-label the easy noisy samples in $\mathcal{U}'_E$ and place them together with the clean samples in $\mathcal{X}'$ for the MixUp data augmentation with $\mathcal{D}'=\mathcal{X}'\cup \mathcal{U}'_E$.  This operation enables us to obtain the proportionally mixed set $\hat{\mathcal{D}}=MixUp(\mathcal{D}',\text{shuffled}(\mathcal{D}'))$ where the proportion of samples from the noisy set, denoted by $\zeta=\frac{|\mathcal{U}'_E|}{|\mathcal{X}'|+|\mathcal{U}'_E|}$, will be larger than $0.5$ for high noise rate problems. 

To optimise the classification term we rely on the regularised CE loss~\citep{DivideMix}:
\begin{equation}
    \mathcal{L}= \ell_{CE}  +\lambda_r\ell_{r},
    \label{eq:max_likelihood_loss}
\end{equation}
with $\ell_{CE} = -\frac{1}{|\mathcal{\hat{D}}|}\sum_{(\tilde{\mathbf{x}},\tilde{\mathbf{y}})\in\mathcal{\hat{D}}} \tilde{\mathbf{y}}^{\top}\log p_{\theta}(. | f_{\phi}(\tilde{\mathbf{x}}))$ 
 and 
$ \ell_r = KL \left [ \pi_{|\mathcal{Y}|} \Bigg | \Bigg | \frac{1}{|\mathcal{\hat{D}}| } \sum_{ (\mathbf{x},\mathbf{y}) \in \mathcal{D}'} p_{\theta}(.|f_{\phi}(\mathbf{x})) \right ]$,  
where $\lambda_{r}$ weights the regularisation loss, and $\pi_{|\mathcal{Y}|}$ denotes a vector of $|\mathcal{Y}|$ dimensions with values equal to $1/|\mathcal{Y}|$. 
Different from SOTA semi-supervised methods~\citep{DivideMix,AugDesc}, we do not have an additional loss for the noisy set because we assume that most of the samples in the noisy set have correct predictions (this is empirically shown in Sec.~\ref{sec:analysis}). 
A positive outcome from not using such additional loss for the noisy set is that we no longer need to manually set hyper-parameter values that depend on the noise rate of the problem, as is done by DivideMix~\citep{DivideMix}.
%SOTA noise-robust classifiers~\citep{DivideMix,nguyen2019self, elr2020} are formed by an ensemble of two classifiers, where the classifier structure is the same, but their parameters are denoted by $\theta(1),\theta(2) \in \Theta$.  The training for $\theta(1)$ influences $\theta(2)$ and vice-versa, where this can be achieved by co-training~\citep{DivideMix, elr2020}  or student-teacher~\citep{nguyen2019self} approaches. Our training relies on co-training. The estimation of class prediction for label estimation and evaluation are given by the average outputs of the models. 
The pseudo-code for the training of PropMix is shown in Algorithm 1 in the supplementary material.

\vspace{-.1in}
\section{Experiments}

\subsection{Data sets}

We conduct our experiments on the data sets  CIFAR-10, CIFAR-100~\citep{krizhevsky2009learning}, Controlled Noisy Web Labels (CNWL)~\citep{jiang2020beyond}, Clothing1M~\citep{xiao2015learning}  and  WebVision~\citep{webvision}. CIFAR-10 and CIFAR-100 have 50k training and 10k testing images of size $32 \times 32$ pixels, where CIFAR-10 has 10 classes and CIFAR-100 has 100 classes and all training and testing sets have equal number of images per classes. 
As CIFAR-10 and CIFAR-100 data sets originally do not contain label noise, we follow the literature~\citep{DivideMix} and add  the following synthetic noise types (see Sec.~\ref{sec:problem_definition}): symmetric (with noise rate $\eta \in \{0.2, 0.5, 0.8, 0.9\}$, as defined in Sec.~\ref{sec:problem_definition}), asymmetric (using the mapping  in~\citep{DivideMix, patrini2017making}, with $\eta_{jc}=0.4$), and semantic~\citep{rog} (with noisy labels based on a trained VGG~\citep{vgg}, DenseNet~(DN), and ResNet~(RN)). 

 The CNWL dataset~\citep{jiang2020beyond} is a benchmark to study real-world web label noise in a controlled setting. Both images and labels are crawled from the web and the noisy labels are determined by matching images. The controlled setting provide different magnitudes of label corruption in real applications, varying from 0\% to 80\%. %CNWL provides controlled web noise for Mini-ImageNet dataset, called red noise. The 
 We study the red Mini-ImageNet that consists of 50k training images and 5k test images, with 100 classes. 
 The original 84$\times$84-pixel images are resized to 32$\times$32 pixels. 
 The noise rates are 20\%, 60\% and 80\%, as used in~\citep{FaMUS}.
 
 Clothing1M consists of 1 million training images acquired from online shopping websites and it has 14 classes.
As the images from the data set vary in size, we resized the images to $256 \times 256$ for training, as used in \cite{DivideMix,han2019deep}.
The data set provide additional clean sets for 
training, validation, and testing of 50k, 14k and 10k images, respectively. For our experiments we do not use any of the clean training or validation sets, but we use the clean test set for evaluation.
 
 WebVision contains 2.4 million images collected from the internet, with the same 1000 classes from ILSVRC12~\citep{deng2009imagenet} and images resized  to $256 \times 256$ pixels. It provides a clean test set of 50k images, with 50 images per class. We compare our model using the first 50 classes of the Google image subset, as used in \cite{DivideMix, chen2019understanding}.
% , which reduces the training set and test size to

% Clothing1M consists of 1 million training images acquired from online shopping websites and it is composed of 14 classes. As the images from the data set vary in size, we resized the images to $256 \times 256$ for training, as used in \cite{li2020dividemix, han2019deep}. The data set is heavily imbalanced and most of the noise is asymmetric~\citep{yi2019probabilistic}, with noise rate estimated to be around 40\%~\citep{xiao2015learning}. The data set provide additional clean sets for training, validation, and test of 50k, 14k and 10k images, respectively. For our experiments we do not use any of the clean training or validation sets, but we use the test set for evaluation.

\subsection{Implementation}

For CIFAR-10 and CIFAR-100 we used a 18-layer 
PreaAct-ResNet-18 (PRN18)~\citep{he2016identity}  
as our backbone model~\citep{DivideMix}. The models are trained with stochastic gradient descent (SGD) with momentum of 0.9, weight decay of 0.0005 and batch size of 64. For the self-supervised pre-training learning task, we adopt SimCLR~\citep{SimCLR} with a batch size of 1024, SGD optimiser with a learning rate of 0.4, decay rate of 0.1, momentum of 0.9 and weight decay of 0.0001, and run it for 800 epochs.  This pre-trained model produces feature representations of 128 dimensions. Using these representations we mine $K = 20$ nearest neighbours (as in \citep{SCAN}) for each sample to form the sets $\{ \mathcal{N}_{\mathbf{x}_{i}}\}_{i=1}^{|\mathcal{D}|}$, defined in Sec.~\ref{sec:propmix}. In the supervised training stage, the model is trained with SGD with momentum of 0.9, weight decay of 0.0005 and batch size of 64. The learning rate is 0.02 which is reduced to 0.002 in the middle of the training. The WarmUp and total number of epochs is defined according to each data set, as defined in~\citep{DivideMix}. 
For CIFAR-10 and CIFAR-100, PRN18 is trained with a WarmUp stage of 30 epochs for CIFAR-10 and 10 epochs for CIFAR-100, and 300 epochs of final training. In our training, we also use a co-teaching approach, as in~\citep{DivideMix, elr2020, han2018co}. 
We estimate noise rate with $|\mathcal{U}|/|\mathcal{D}|$, defined in Eq.~\ref{eq:clean_noisy_sets} -- if this ratio is larger than 50\%, we use strong data augmentation with cutout~\citep{cubuk2019autoaugment}, otherwise, we use standard augmentation (i.e., crop and flip). %\gustavo{We should say why we do this, or cite anyone.}. 

For the semantic noise from~\citep{rog}, we use  DenseNet-100~\citep{iandola2014densenet} as backbone, following~\citep{rog}. The pre-training stage uses the same parameters as in CIFAR, except that we use a batch size of 512. In the supervised stage, we use the same protocol as~\citep{rog}, which uses SGD with momentum 0.9, weight decay 10-4 and learning rate of 0.1 that is divided by 10 after epochs 40 and 80 for CIFAR-10 (which runs for 120 epochs in total), and after epochs 80, 120 and 160 for CIFAR-100 (runs for 200 epochs in total). The WarmUp epochs is the same as CIFAR-10/CIFAR-100.

For red Mini-Imagenet we use PRN18 as backbone, following~\citep{FaMUS}. For the self-supervised pre-training task, we adopt SimCLR~\citep{SimCLR} with batch size 128. All other parameters for the self-supervised pre-training are the same as described for CIFAR. For the supervised stage, we adopt the implementation of~\citep{FaMUS}, where we train for 300 epochs,  relying on SGD with learning rate of 0.02 (decreased by a factor of ten at epochs 200 and 250), momentum of 0.9 and weight decay of 5e-4. %We also resized the images from $84\times84$ to $32\times32$~\citep{FaMUS}.

For Clothing1M, we use ResNet-50 as backbone, following ~\citep{DivideMix}. In this protocol, a ResNet-50 with ImageNet~\citep{deng2009imagenet} pre-trained weights is used and we decided to not use the self-supervised stage for this experiment because the model is already pre-trained. The ResNet-50 is trained for 80 epochs, including a WarmUp stage of 1 epoch, with a batch size of 32, SGD with a learning rate of 0.002 (divided by 10 at epoch 40), momentum of 0.9 and weight decay of 0.0001.

For Webvision, we use InceptionResNet-V2 as backbone, following~\citep{DivideMix}. For self-supervised pre-training, we adopt MoCo-v2 (4-GPU training)~\citep{MoCoV2}, trained with 100 epochs, with a batch size of 128, SGD with a learning rate of 0.015 (divided by 10 at epoch 50), momentum of 0.9 and weight decay of 0.0001, and run it for 100 epochs with a WarmUp stage of 1 epoch. The feature representations learned from this process have 128 dimensions. All the other parameters were the same as described above for CIFAR. 
%The number of epochs $E= 100$, and the learning rate is divided by 10 at epoch 50.

% For the Clothing1M, we use the same setup as \cite{DivideMix}, which used a ResNet-50 pretrained on ImageNet. As this setup already uses a pretrained model, we didn't use any self-supervised training. The model is training for 80 epochs. The learning rate is is 0.01 and it is decreased by a factor of ten at epoch 40.

\subsection{PropMix Analysis}
\label{sec:analysis} 

\begin{figure}[!ht]
\centering
\includegraphics[width=0.6\columnwidth]{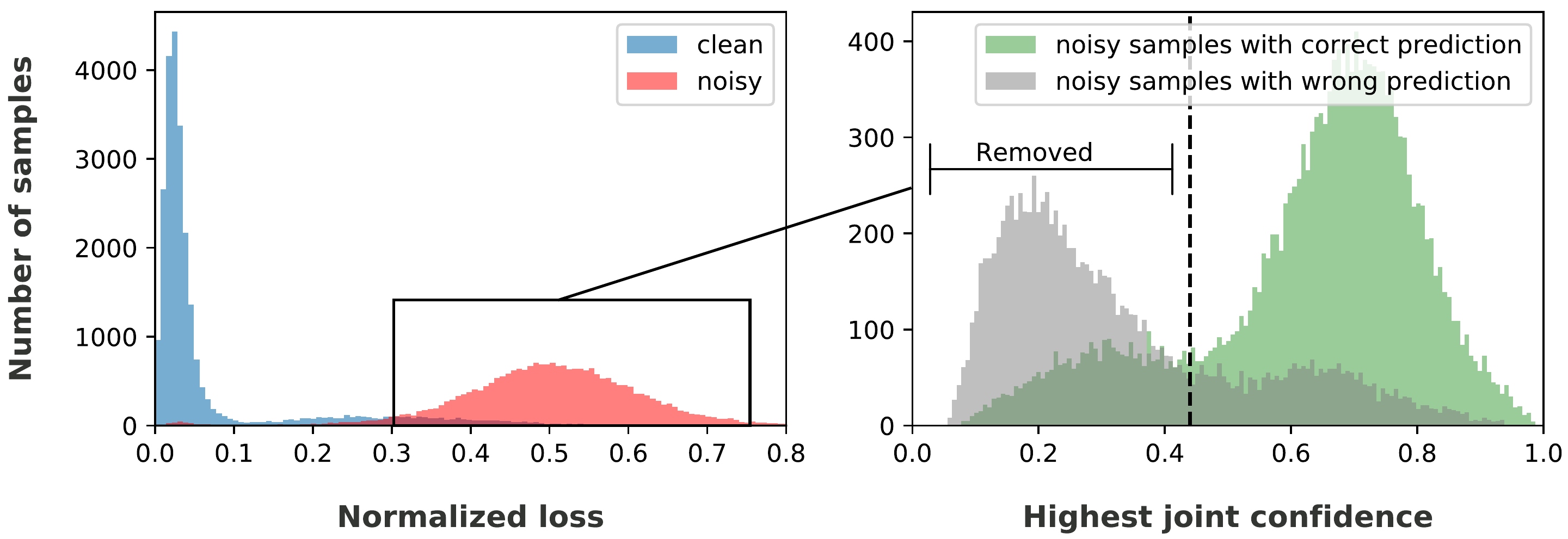}
% \vspace{0.1in}
\caption{PropMix classification of easy and hard noisy samples. From the (normalized) loss of all samples, PropMix uses GMM to split clean and noisy samples, and then, using the largest classification confidence, it uses another GMM to split the hard and easy noisy samples. This example was run on CIFAR-100, 80\% noise rate at epoch 150.  }
\label{fig:propmix_hist}
\end{figure}

\begin{figure}[!ht]
\centering
\includegraphics[width=0.92\columnwidth]{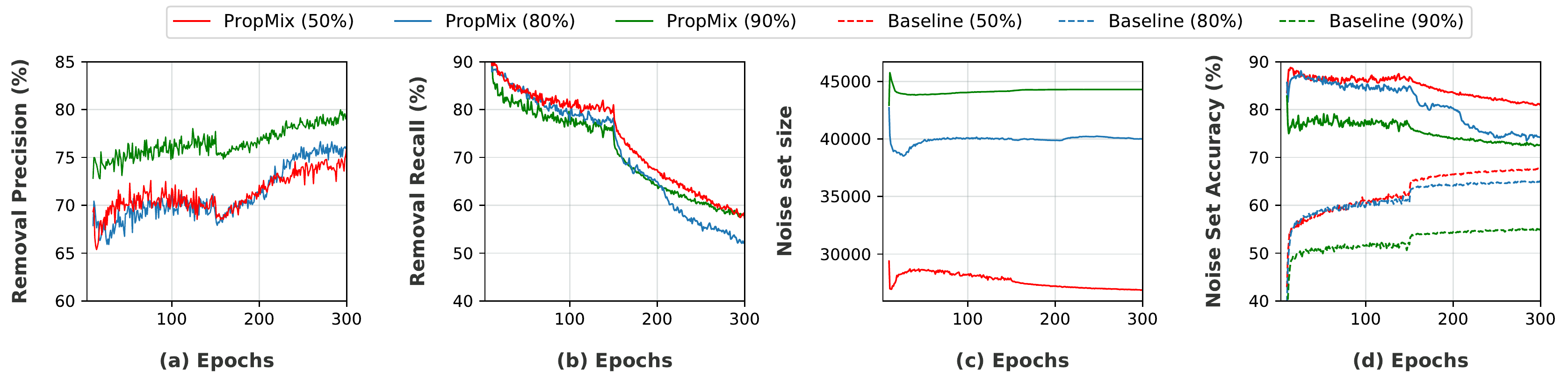}
% \vspace{0.1in}
\caption{Evaluation of hard noisy sample filtering, for CIFAR-100, with 50\% to 90\% symmetric noise. Graphs in (a,b) show the precision and recall of the classification of hard noisy samples (i.e., noisy samples that are incorrectly labelled by the model), Graph~(c) shows the estimated noise set size, and (d) shows the classification accuracy of the re-labelled easy noisy samples, compared to the baseline that does not filter out hard noisy samples.}
\label{fig:noisy_pred}
\end{figure}

\begin{figure}[!ht]
\centering
\includegraphics[width=0.90\columnwidth]{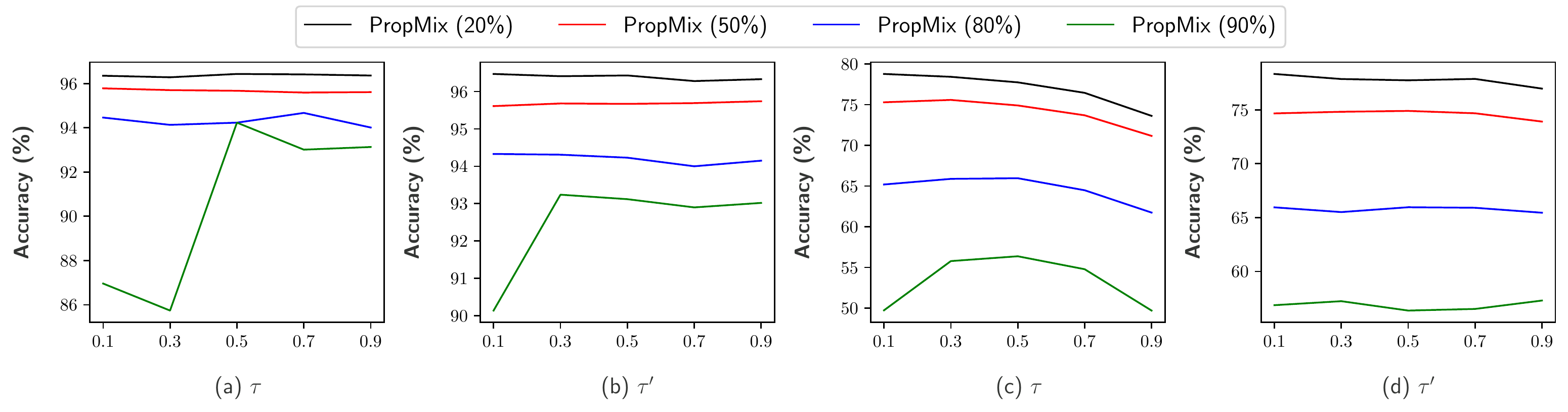}
% \vspace{0.1in}
\caption{Evaluation of the parameters $\tau$~(clean sample threshold) and $\tau'$~(hard noisy sample threshold), for CIFAR-10 and CIFAR-100, with 20\% to 90\% symmetric noise. Graphs in (a,b) show the best accuracy when training using PropMix with CIFAR-10, and graphs (c,d) show results for CIFAR-100, at different values of $\tau$ and $\tau'$. Graph (a,c) vary the parameter $\tau$, while fixes $\tau'$ at 0.5. Graph (b,d) vary the parameter $\tau'$ while fixes $\tau$ at 0.5.}
\label{fig:th}
\end{figure}

 Fig.~\ref{fig:propmix_hist} shows the PropMix hard noisy sample filtering process for CIFAR-100 with 80\% symmetric noise, at epoch 150 (i.e., half of the training epochs). 
 The left histogram shows the clean and noisy classification from Eq.~\ref{eq:clean_noisy_sets}, while the right one shows the easy and hard noisy sample classification from Eq.~\eqref{eq:hard_noisy_set}.
As can be seen, the use of confidence in Eq.~\eqref{eq:hard_noisy_set}, instead of the loss from Eq.~\ref{eq:clean_noisy_sets}, is an effective way to remove noisy samples with wrong prediction. 
We also evaluated the quality of the filtering stage using PropMix for CIFAR-100 under different symmetric noise rates. Fig.~\ref{fig:noisy_pred}(a) and (b) show the precision and recall of the hard noisy sample filtering as a function of training epochs.  These graphs show that the larger the noise, the higher the precision. 
For example, for 90\% noise, the hard noise classification reaches 80\% at the end of training. 
Fig.~\ref{fig:noisy_pred}(c) shows the size of the  noisy sample set during training. 
Fig. \ref{fig:noisy_pred}(d) compares the easy noise set re-labelling accuracy by PropMix with a baseline that does not filter out hard noisy samples. 
We can see that PropMix is substantially more accurate,
which enables a more successful training with the easy noisy samples. We also evaluated the impact of the parameters $\tau$ and $\tau'$ in the training. Fig.~\ref{fig:th} shows the PropMix accuracy for CIFAR-10 (a-b) and CIFAR-100 (c-d) under different symmetric noise rates, varying the parameters $\tau$, which is related to the clean/noisy filtering from Eq.~\ref{eq:clean_noisy_sets}, and $\tau'$, which related to hard noise filtering, from Eq.~\ref{eq:hard_noisy_set}. Fig.~\ref{fig:th}(a,c) vary $\tau$ while fixing $\tau'=0.5$, and Fig.~\ref{fig:th}(b,d) vary $\tau'$ while fixing $\tau=0.5$. We can see that $\tau'$ does not have a strong impact on the accuracy, which motivated us to use $\tau'=0.5$ independently of the noise rate and data set. The parameter $\tau$ is shown to decrease accuracy at higher values. In PropMix we use $\tau=0.5$ as in other works~\cite{DivideMix}. 

% \begin{figure}[!ht]
% \centering
% \includegraphics[width=0.5\columnwidth]{images/proportion.pdf}
% \caption{Proportion of samples from predicted clean and noisy sets during MixUp operations, for the PropMix, compared to baseline using MixMatch. }
% \label{fig:noisy_pred}
% \end{figure}

\vspace{-.1in}
\subsection{Comparison with State-of-the-Art}

For CIFAR-10 and CIFAR-100, we evaluate our model using symmetric label noise ranging from 20\% to 90\%, and 40\% asymmetric noise. 
We report both the best test accuracy across all epochs and the averaged test accuracy over the last 10 epochs of training, similar to~\citep{DivideMix}. Tab.~\ref{tab:results_cifar} shows that for CIFAR-10 and CIFAR-100 data sets, our method obtains the best results for almost all evaluated noisy rates.

\begin{table*}[t!]
\centering
%\scriptsize
% \footnotesize
\scalebox{0.6}{
% \begin{tabular}{p{2cm}|p{0.2cm}p{0.2cm}p{0.2cm}p{0.3cm}|p{0.4cm}||p{0.2cm}p{0.2cm}p{0.2cm}p{0.2cm}}
\begin{tabular}{lc|cccc|c||cccc}
\hline
%\toprule
\multicolumn{2}{c}{dataset} & \multicolumn{5}{c}{CIFAR-10} & \multicolumn{4}{c}{CIFAR-100}\\    
% \cmidrule(lr){2-5} \cmidrule(lr){6-9} 
%\midrule
\hline
\multicolumn{2}{c}{Noise type} & \multicolumn{4}{c}{sym.} & \multicolumn{1}{c}{asym.} &  \multicolumn{4}{c}{sym.} \\
%\midrule
\hline
Method/ noise ratio & &  20\% & 50\% & 80\% & 90\% & 40\% & 20\% & 50\% & 80\% & 90\% \\
%\midrule
\hline
\multirow{2}{*}{Cross-Entropy \citep{DivideMix}}& Best & 86.8 & 79.4 & 62.9 & 42.7 & 85.0 &  62.0 & 46.7 & 19.9 & 10.1\\
    & Last & 82.7 & 57.9 & 26.1 & 16.8 & 72.3 &  61.8 & 37.3 & 8.8 & 3.5\\
%\midrule
\hline
% Bootstrap \citep{reed2014training} & 86.8 & 79.8 & 63.3& 42.9& - & 62.1& 46.6 & 19.9 & 10.2\\
\multirow{2}{*}{Coteaching+ \citep{yu2019does}} & Best & 89.5 & 85.7 & 67.4& 47.9& - &65.6& 51.8 & 27.9 & 13.7\\
    & Last & 88.2 & 84.1 & 45.5& 30.1& - & 64.1& 45.3 & 15.5 & 8.8\\
%\midrule
\hline
\multirow{2}{*}{MixUp \citep{mixup}} &Best & \textbf{95.6} & 87.1 & 71.6& 52.2& - & 67.8& 57.3 & 30.8 & 14.6\\
 &Last & 92.3 & 77.3 & 46.7& 43.9& - &  66.0& 46.6 & 17.6 & 8.1\\
%\midrule
\hline
\multirow{2}{*}{PENCIL \citep{yi2019probabilistic}}& Best & 92.4 & 89.1 & 77.5& 58.9& 88.5 &  69.4& 57.5 & 31.1 & 15.3 \\
  & Last & 92.0 & 88.7 & 76.1& 58.2& 88.1 &  68.1& 56.4 & 20.7 & 8.8 \\
%\midrule
\hline
\multirow{2}{*}{Meta-Learning \citep{li2019learning}}&Best & 92.9 & 89.3 & 77.4& 58.7& 89.2 &  68.5& 59.2 & 42.4 & 19.5 \\
   &Last & 92.0 & 88.8 & 76.1& 58.3& 88.6 &  67.7& 58.0 & 40.1 & 14.3 \\
%\midrule
\hline
\multirow{2}{*}{M-correction \citep{arazo2019unsupervised}}&Best& 94.0 & 92.0 & 86.8& 69.1& 87.4 &  73.9& 66.1 & 48.2 & 24.3 \\
   &Last& 93.8 & 91.9 & 86.6& 68.7& 86.3 &  73.4& 65.4 & 47.6 & 20.5 \\
%\midrule
\hline
\multirow{2}{*}{MentorMix~\citep{jiang2020beyond}}& Best & \textbf{95.6} & - & 81.0 & - & - &  \textbf{78.6} & - & 41.2 & - \\
  & Last & - & - & - & - &  - & - & - & - & - \\
%\midrule
\hline
\multirow{2}{*}{MOIT+~\citep{ortego2020multi}}& Best & 94.1 & - & 75.8 & - & 93.3 &  75.9 & - & 51.4 & - \\
  & Last & - & - & - & - & - &  - & - & - & - \\
%\midrule
\hline
\multirow{2}{*}{DivideMix \citep{DivideMix}}& Best & \textbf{96.1} & 94.6 & \textbf{93.2} & 76.0& 93.4  & 77.3 & \textbf{74.6} & 60.2 & 31.5 \\
  & Last & \textbf{95.7} & 94.4 & \textbf{92.9} & 75.4& 92.1  & \textbf{76.9} & \textbf{74.2} & 59.6 & 31.0 \\
%\midrule
\hline
\multirow{2}{*}{ELR+~\citep{liu2020early}}& Best & \textbf{95.8} & \textbf{94.8} & \textbf{93.3} & 78.7& 93.0 &  \textbf{77.6} & 73.6 & 60.8 & 33.4 \\
  & Last & - & - & - & - & - &  - & - & - & - \\
%\midrule
\hline
% \multirow{2}{*}{DM-AugDesc-WS-WAW~\citep{AugDesc}}& Best & \textbf{96.3} & \textbf{95.4} & \textbf{93.8} & \textbf{91.9} & \textbf{94.6} & - & \textbf{79.5} & \textbf{77.2} & 
% \textbf{66.4} & 41.2 \\
%   & Last & 96.2 & 95.1 & 93.3 & \textbf{91.8} & \textbf{94.3} & - & \textbf{79.2} & \textbf{77.0} & \textbf{66.1} & 40.9 \\
% \midrule

\multirow{2}{*}{\textbf{PropMix (Ours)}}& Best & \textbf{96.44} & \textbf{95.77} & \textbf{93.94} & \textbf{93.48} & \textbf{94.89} &  77.41 & \textbf{74.56} & \textbf{67.34} & \textbf{58.57}\\
    & Last & \textbf{96.09} & \textbf{95.53} & \textbf{93.77} & \textbf{93.20} & \textbf{94.64} &  \textbf{76.99} & \textbf{73.71} & \textbf{66.75} & \textbf{58.32}\\
%\bottomrule \\
\hline
% \multirow{2}{*}{Improvement}& Best & - & - & - & - & - & - & - & - & - & -\\
%   & Last & - & - & - & - & - & - & - & - & - & -\\
% \bottomrule \\
\end{tabular}
}
% \vspace{0.1in}
\caption{Test accuracy (\%) for all competing methods on CIFAR-10 and CIFAR-100 under symmetric and asymmetric noises. Results from related approaches are as presented in~\citep{DivideMix}. 
%The results with (*) were produced by locally running the published code provided by the authors. 
Top methods within $1\%$ are in \textbf{bold}.} 
\label{tab:results_cifar}
\end{table*}

%We also evaluate our method on semantic noise, which is harder and more realistic than the synthetic noise.
Tab.~\ref{tab:res_semantic} shows the results for semantic noise~\citep{rog}, which is harder and more realistic than the synthetic noise, where PropMix shows significantly more accurate results than any of the methods in~\citep{rog}. 
In Tab.~\ref{tab:results_red_noise}, we show that PropMix improves the SOTA by a large margin on the Red Mini-ImageNet semantic noise experiment from~\citep{jiang2020beyond}.

% Results show that ...\textcolor{red}{[waiting for results ...]}. Tab.~\ref{tab:results_red_noise} shows the Red Mini-ImageNet semantic noise experiment from~\citep{jiang2020beyond}, with results extracted from~\citep{jiang2020beyond}. The results show that our PropMix improves the SOTA by a large margin. 
%We believe that because our method remove the hard noisy samples, which have semantic noise, we can provide a cleaner training data, which improves the model generalization.

\begin{table*}[ht]
\centering
%\scriptsize
% \footnotesize
\scalebox{0.65}{
%\begin{tabular}{@{}p{2.7cm}@{}p{0.5cm}p{0.5cm}p{0.6cm}p{0.5cm}p{0.5cm}p{0.5cm}}
\begin{tabular}{ccccccc}
\toprule
Data set & \multicolumn{3}{c}{CIFAR-10} & \multicolumn{3}{c}{CIFAR-100}\\    
% \cmidrule(lr){2-5} \cmidrule(lr){6-9} 
% \midrule
% \multicolumn{2}{c}{Noise type} & \multicolumn{4}{c}{sym.} & \multicolumn{2}{c}{asym.} &  \multicolumn{4}{c}{sym.} \\
\midrule

Method/ noise ratio & DenseNet (32\%) & ResNet (38\%) & VGG (34\%) & DenseNet (34\%) & ResNet (37\%) & VGG (37\%) \\
\midrule
D2L + RoG~\citep{lee2019robust} &  68.57 & 60.25 & 59.94 & 31.67 & 39.92 & 45.42\\
CE + RoG~\citep{lee2019robust} &  68.33 & 64.15 & 70.04 & 61.14 & 53.09 & 53.64\\
Bootstrap + RoG~\citep{lee2019robust} &  68.38 & 64.03 & 70.11 & 54.71 & 53.30 & 53.76\\
Forward + RoG~\citep{lee2019robust} &  68.20 & 64.24 & 70.09 & 53.91 & 53.36 & 53.63\\
Backward + RoG~\citep{lee2019robust} &  68.66 & 63.45 & 70.18 & 54.01 & 53.03 & 53.50\\
% DivideMix* \cite{DivideMix} &  82.25 & 80.42 & 84.14 & \textbf{60.50} & \textbf{61.02} & \textbf{61.83}\\
\textbf{PropMix (Ours)} &  \textbf{84.25} & \textbf{82.51} & \textbf{85.74} & \textbf{60.98} & \textbf{58.44} & \textbf{60.01} \\
\bottomrule \\
\end{tabular}
}
\caption{Test accuracy (\%) for the semantic noise benchmark~\citep{lee2019robust}, where baseline results are from~\citep{lee2019robust}. The results with (*) were produced by locally running the published code provided by the authors. Top methods ($\pm1\%$) are in \textbf{bold}.} 
\label{tab:res_semantic}
\end{table*}

% \begin{table*}
% \centering
% %\scriptsize
% \scalebox{0.7}{
% \begin{tabular}{lcccc}
% \toprule
% Method/ noise ratio & 20\% & 40\% & 60\%& 80\% \\
% \midrule
%  Cross-entropy~\citep{FaMUS}    & 47.36 & 42.70 & 37.30 & 29.76 \\
%  MixUp~\citep{mixup}          & 49.10 & 46.40 & 40.58 & 33.58 \\
%  DivideMix~\citep{DivideMix}     & 50.96 & 46.72 & 43.14 & 34.50 \\
% % DivdeMix(*)   & 51.06    &  53.20  &   48.54   &  36.54  \\
%  MentorMix~\citep{jiang2020beyond}  & 51.02 & 47.14 & 43.80 & 33.46 \\
%  FaMUS~\citep{FaMUS}  & 51.42 & 48.06 & 45.10 & 35.50 \\
%  \textbf{PropMix (Ours)} & \textbf{61.24} & \textbf{56.22} & \textbf{52.84} & \textbf{43.42} \\
% \bottomrule \\
% \end{tabular}
% }
% \caption{Test accuracy (\%) for Red Mini-ImageNet~\citep{jiang2020beyond}. Results from baseline methods are as presented in \citep{FaMUS}. Top methods ($\pm1\%$) are in \textbf{bold}.}
% \label{tab:results_red_noise}
% \end{table*}

We also evaluate PropMix on the noisy large-scale datasets WebVision~\citep{webvision} and Clothing1M~\citep{xiao2015learning}. Tab.~\ref{tab:results_clothing} shows competitive results compared to SOTA. As Clothing1M experimental protocol uses ImageNet pre-trained weights, the training could not benefit from PropMix strategy. Tab.~\ref{tab:res_WebVision} shows the Top-1/-5 test accuracy using the WebVision and ILSVRC12 test sets. Results show that PropMix is slightly better than the SOTA for WebVision test set. This suggests that our approach is also effective in large-scale, low noise rate problems.

% We also evaluate PropMix on the noisy large-scale dataset Clothing!M. Table \ref{res_clothing} shows the best accuracy using the Clothing1M. \textcolor{red}{[complete...]}

% \begin{table}[t]
% % \small
% % \scriptsize
% \footnotesize
% \centering
% \scalebox{0.95}{
% \begin{tabular}{lcc}
% \toprule
% Method & Test Accuracy \\
% \midrule
%  Cross-Entropy~\citep{li2020dividemix}  & 69.21  \\
% %  Cross-Entropy  & 69.21  \\
% %  F-correction   & 69.84 \\
%  M-correction \cite{arazo2019unsupervised}   & 71.00 \\
% %  M-correction  & 71.00 \\
% %  Joint-Optim.~\citep{tanaka2018joint}      & 72.16  \\
% %  Meta-Cleaner~\citep{zhang2019metacleaner}    & 72.50  \\
% %  Meta-Learning~\citep{li2019learning}   & 73.47 \\
%  PENCIL\cite{yi2019probabilistic} &   73.49 \\
% %  PENCIL &   73.49 \\
%  DeepSelf~\citep{han2019deep} & 74.45 \\
% %  DeepSelf & 74.45 \\
%  CleanNet~\citep{lee2018cleannet}    & 74.69 \\
% %  CleanNet   & 74.69 \\
%  DivideMix~\citep{li2020dividemix}      & \textbf{74.76} \\
% %  DivideMix     & \textbf{74.76} \\
%  PropMix~$\dagger$~[ours]      &  73.81 \\
% %  Zhang \cite{zhang2020distilling} & 80.0 \\

%  \bottomrule
%  \\
% \end{tabular}
% }
% \caption{Results for Clothing1M~\citep{xiao2015learning}. Results from baseline methods are as presented in \cite{li2020dividemix}. The marker $\dagger$ denotes the model is trained from scratch.}
% \label{tab:res_clothing}
% \end{table}

\subsection{Ablation Study}
\label{sec:ablation}

We show the ablation study of PropMix with CIFAR-10 and CIFAR-100 under symmetric and asymmetric noises at several rates. In Tab.~\ref{tab:results_ablation}, we show \emph{Self-superv. pre-train}, which has the result from self-supervised training (with SCAN~\citep{SCAN}), where the result is the same across different noise rates because it never uses the noisy labels for training. 
\emph{SSL (DivideMix)}~\citep{DivideMix}) is current SOTA in noisy label learning, but it has worse accuracy in high noise rate problems (> 80\% noise) than SCAN. \emph{Self-superv. pre-train+SSL (DivideMix)} pre-trains DivideMix with SCAN to improve accuracy in high-noise rate problems, without affecting low-noise rate results.
Just removing the MSE loss in \emph{Self-superv. pre-train+SSL (DivideMix) w/o MSE} does not help much because the noisy samples from DivideMix is not accurately re-labelled.
Adding our hard noisy sample filtering in \emph{Self-superv. pre-train+SSL (DivideMix) + filtering with MSE} does not improve accuracy because DivideMix still uses MSE loss to train the easy noisy samples, which is unnecessary given that these samples are  accurately re-labelled and can be trained with the CE loss.
\emph{PropMix} can achieve better results for high noise rates, using a simpler loss function (uses only CE loss) with less hyper-parameters, and removing hard noisy samples.

% %\begin{wraptable}{r}{5.5cm}
% \begin{table}[t!]
% % \scriptsize
% % \footnotesize
% \centering
% \scalebox{0.7}{
% \begin{tabular}{lcccc}
% \toprule
%  dataset & \multicolumn{2}{c}{WebVision} &  \multicolumn{2}{c}{ILSVRC12} \\
% \midrule
% {Method} & {Top-1} & {Top-5} & {Top-1} & {Top-5} \\
% \midrule
%  F-correction~\citep{patrini2017making}    & 61.12 & 82.68 & 57.36 & 82.36 \\
%  Decoupling~\citep{malach2017decoupling}     & 62.54 & 84.74 & 58.26 & 82.26\\
%  D2L~\citep{ma2018dimensionality}            & 62.68 & 84.00  & 57.80 & 81.36\\
%  MentorNet~\citep{jiang2018mentornet}      & 63.00  & 81.40 & 57.80 & 79.92 \\
%  Co-teaching~\citep{han2018co}    & 63.58 & 85.20 & 61.48 & 84.70 \\
%  Iterative-CV~\citep{chen2019understanding}   & 65.24 & 85.34 & 61.60 & 84.98\\
%  MentorMix~\citep{jiang2020beyond}& 76.00 & 90.20 &  72.90 & \textbf{91.10} \\
% %  MOIT+~\citep{ortego2020multi}& 78.76 & - & - & - \\
%  DivideMix~\citep{DivideMix}      & 77.32  & \textbf{91.64} & \textbf{75.20} & \textbf{90.84}\\
%  ELR+~\citep{elr2020}& 77.78 & \textbf{91.68} & 70.29 & 89.76 \\
%  %FaMUS~\citep{FaMUS}*& \textbf{79.40} & \textbf{92.80} & \textbf{77.00} & \textbf{92.76}\\
%  \textbf{PropMix~(Ours)}     & \textbf{78.84} & 90.56 & \textbf{74.64??} & 89.36?? \\
%  \bottomrule \\
% \end{tabular}
% }
% \caption{Test accuracy (\%) for WebVision~\citep{webvision}  by methods trained with 100 epochs. Baseline results are as presented in \cite{DivideMix}. Top methods within $1\%$ are in \textbf{bold}.}
% \label{tab:res_WebVision}
% \end{table}
% %\end{wraptable}
\begin{wraptable}{r}{5.35cm}
%\begin{table}[t!]
% \scriptsize
% \footnotesize
\centering
\scalebox{0.7}{
\begin{tabular}{lcc}
\toprule
{Method} & {Top-1} & {Top-5} \\ 
\midrule
 F-correction~\citep{patrini2017making}    & 61.12 & 82.68  \\
 Decoupling~\citep{malach2017decoupling}     & 62.54 & 84.74 \\
 D2L~\citep{ma2018dimensionality}            & 62.68 & 84.00  \\
 MentorNet~\citep{jiang2018mentornet}      & 63.00  & 81.40  \\
 Co-teaching~\citep{han2018co}    & 63.58 & 85.20  \\
 Iterative-CV~\citep{chen2019understanding}   & 65.24 & 85.34 \\
 MentorMix~\citep{jiang2020beyond}& 76.00 & 90.20  \\
%  MOIT+~\citep{ortego2020multi}& 78.76 & - & - & - \\
 DivideMix~\citep{DivideMix}      & 77.32  & \textbf{91.64} \\
 ELR+~\citep{elr2020}& 77.78 & \textbf{91.68} \\
 %FaMUS~\citep{FaMUS}*& \textbf{79.40} & \textbf{92.80} & \textbf{77.00} & \textbf{92.76}\\
 \textbf{PropMix~(Ours)}     & \textbf{78.84} & 90.56  \\
 \bottomrule \\
\end{tabular}
}
\caption{Test accuracy (\%) for WebVision~\citep{webvision}  by methods trained with 100 epochs. Baselines come from~\cite{DivideMix}. Top methods within $1\%$ in \textbf{bold}.}
\label{tab:res_WebVision}
%\end{table}
\end{wraptable}

\begin{table}
\begin{minipage}{.44\columnwidth}
\centering
\scalebox{0.7}{
\begin{tabular}{lcccc}
\toprule
Method/ noise ratio & 20\% & 40\% & 60\%& 80\% \\
\midrule
 Cross-entropy~\citep{FaMUS}    & 47.36 & 42.70 & 37.30 & 29.76 \\
 MixUp~\citep{mixup}          & 49.10 & 46.40 & 40.58 & 33.58 \\
 DivideMix~\citep{DivideMix}     & 50.96 & 46.72 & 43.14 & 34.50 \\
% DivdeMix(*)   & 51.06    &  53.20  &   48.54   &  36.54  \\
 MentorMix~\citep{jiang2020beyond}  & 51.02 & 47.14 & 43.80 & 33.46 \\
 FaMUS~\citep{FaMUS}  & 51.42 & 48.06 & 45.10 & 35.50 \\
 \textbf{PropMix (Ours)} & \textbf{61.24} & \textbf{56.22} & \textbf{52.84} & \textbf{43.42} \\
\bottomrule \\
\end{tabular}
}
\caption{Test accuracy (\%) for Red Mini-ImageNet~\citep{jiang2020beyond}. Results from baseline methods are as presented in \citep{FaMUS}. Top methods ($\pm1\%$) are in \textbf{bold}.}
\label{tab:results_red_noise}
\end{minipage}
\hspace{0.3in}
\begin{minipage}{.5\columnwidth}
\centering
\scalebox{0.7}{
\begin{tabular}{lcc}
\toprule
Method & Test Accuracy \\
\midrule
 Cross-Entropy~\cite{DivideMix}  & 69.21  \\
%  Cross-Entropy  & 69.21  \\
%  F-correction   & 69.84 \\
 M-correction \cite{arazo2019unsupervised}   & 71.00 \\
%  M-correction  & 71.00 \\
%  Joint-Optim.~\cite{tanaka2018joint}      & 72.16  \\
%  Meta-Cleaner~\cite{zhang2019metacleaner}    & 72.50  \\
%  Meta-Learning~\cite{li2019learning}   & 73.47 \\
 PENCIL\cite{yi2019probabilistic} &   73.49 \\
%  PENCIL &   73.49 \\
 DeepSelf~\cite{han2019deep} & \textbf{74.45} \\
%  DeepSelf & 74.45 \\
 CleanNet~\cite{lee2018cleannet}    & \textbf{74.69} \\
%  CleanNet   & 74.69 \\
 DivideMix~\cite{DivideMix}      & \textbf{74.76} \\
 \textbf{PropMix~(ours)}      &  \textbf{74.30} \\
%  Zhang \cite{zhang2020distilling} & 80.0 \\

 \bottomrule
 \\
 
\end{tabular}

}
\caption{Test accuracy (\%) for Clothing1M~\citep{xiao2015learning}  by methods trained with 80 epochs. Baselines come from~\cite{DivideMix}. Top methods within $1\%$ are in \textbf{bold}.}
\label{tab:results_clothing}
\end{minipage}

% }

\end{table}

% When using SCAN for pre-training DivideMix, we note that results become improves for high noise levels, while have a small improvement to low noise rate. We also evaluated adding the filtering stage to DivideMix, and can see that it doesn't improve the results, in general. This is important to show that we need a different training strategy to leverage the noise filtering, as proposed by PropMix. We also show that using DivideMix without the MSE hurts the training for high noise rate. In PropMix, we show that using MSE is not necessary, and thus we simplify the loss function, reducing the amount of parameters, compared to DivideMix.

\begin{table*}[ht]
\centering
% \scriptsize
\footnotesize
\scalebox{0.75}{
% \begin{tabular}{p{2cm}|p{0.2cm}p{0.2cm}p{0.2cm}p{0.3cm}|p{0.4cm}||p{0.2cm}p{0.2cm}p{0.2cm}p{0.2cm}}
\begin{tabular}{lc|cccc|c||cccc}
\toprule
\multicolumn{2}{c}{dataset} & \multicolumn{5}{c}{CIFAR-10} & \multicolumn{4}{c}{CIFAR-100}\\    
% \cmidrule(lr){2-5} \cmidrule(lr){6-9} 
\midrule
\multicolumn{2}{c}{Noise type} & \multicolumn{4}{c}{sym.} & \multicolumn{1}{c}{asym.} &  \multicolumn{4}{c}{sym.} \\
\midrule

Method/ noise ratio & &  20\% & 50\% & 80\% & 90\% & 40\% & 20\% & 50\% & 80\% & 90\% \\
\midrule

\multirow{1}{*}{Self-superv. pre-train~\citep{SCAN}}& & 81.6 & 81.6  & 81.6 & 81.6 & 81.6 & 44.0 & 44.0 & 44.0 & 44.0 \\

%\multirow{1}{*}{Self-superv. pre-train*~\citep{SCAN}}& & 77.5  & 77.5 & 77.5 & 77.5 & 77.5 & 37.1 & 37.1 & 37.1 & 37.1\\

\multirow{1}{*}{SSL (DivideMix)~\citep{DivideMix}}& & \textbf{96.1} & 94.6 & \textbf{93.2} & 76.0& 93.4  & \textbf{77.3} & \textbf{74.6} & 60.2 & 31.5 \\

\multirow{1}{*}{Self-superv. pre-train + SSL (DivideMix)*}& & 
\textbf{96.2} &
\textbf{94.8} &
\textbf{93.9} &
92.2 &
93.3 &
\textbf{77.4} &
\textbf{74.7} &
\textbf{66.7} &
56.2 \\

\multirow{1}{*}{Self-superv. pre-train + SSL (DivideMix)* w/o MSE}& & 
\textbf{96.2} &
\textbf{95.2} &
91.4 &
85.1 &
93.3 &
\textbf{77.2} &
73.9 &
62.6 &
52.1 \\

\multirow{1}{*}{Self-superv. pre-train + SSL (DivideMix)*} + filtering with MSE & & 
\textbf{96.3} &
\textbf{95.3} &
\textbf{93.8} &
91.6 &
93.0 &
\textbf{77.3} &
\textbf{75.3} &
\textbf{66.7} &
55.7 \\

\multirow{1}{*}{\textbf{PropMix (Ours)}}& & \textbf{96.4} & \textbf{95.8} & \textbf{93.9} & \textbf{93.5} & \textbf{94.9}  & \textbf{77.4} & \textbf{74.6} & \textbf{67.3} & \textbf{58.6}\\
\bottomrule \\
\end{tabular}
}
\caption{In this ablation study we show the test accuracy (\%) on CIFAR-10 and CIFAR-100 under symmetric and asymmetric noise problems.  First, we show the results of self-supervised pre-training with SCAN~\citep{SCAN}. 
Then we show the current SOTA SSL learning for noisy label DivideMix~\citep{DivideMix}.  Next, we show the results of DivideMix pre-trained with SCAN.  
Then, we remove the MSE loss to train DivideMix re-labelled noisy samples, and add the hard noisy filtering stage (but keeping the MSE loss) to the pre-trained DivideMix.
The last row shows our PropMix . The top results within $1\%$ are highlighted.}
\label{tab:results_ablation}
\end{table*}

\vspace{-.3in}
\section{Conclusion and Future Work}
% \vspace{0.1in}

We presented PropMix, a noisy label training algorithm to filter out hard noisy samples, remove samples with incorrect label and incorrect prediction from noisy set with the goal to keep the easy noisy samples that have a higher chance to be correctly re-labelled. PropMix reduces noisy-dependent parameters, while promoting the use of the entire filtered noisy set in a fully supervised training, with proportional MixUp data augmentation on the clean set. Our results on CIFAR-10/100, Red Mini-ImageNet, and WebVision outperform the SOTA methods, also demonstrating robustness to over-fitting in several noise rates with substantial improvement in high noise problems. %compared to SOTA approaches

\section{Acknowledgment}
IR and GC gratefully acknowledge the support of the Australian Research Council through the Centre of Excellence for Robotic Vision CE140100016 and Future Fellowship (to GC) FT190100525.  GC acknowledges the support by the Alexander von Humboldt-Stiftung for the renewed research stay sponsorship. 

\bibliography{egbib}

\newpage

\appendix

\section{PropMix Algorithm}
\label{sec:algorithm}

SOTA noise-robust classifiers~\citep{DivideMix,nguyen2019self, elr2020} are formed by an ensemble of two classifiers, where the classifier structure is the same, but their parameters are denoted by $\theta(1),\theta(2) \in \Theta$.  The training for $\theta(1)$ influences $\theta(2)$ and vice-versa, where this can be achieved by co-training~\cite{DivideMix, elr2020}  or student-teacher~\cite{nguyen2019self} approaches. Our training relies on co-training. The estimation of class prediction for label estimation and evaluation are given by the average outputs of the models. The pseudo-code for the training of PropMix is shown in Algorithm~\ref{alg:PM}.

\begin{algorithm}[h!]
\scriptsize

% % change algorithm font size
% \AlFnt{\footnotesize}
\SetAlgoLined
\KwInput{$\mathcal{D}$, number of epochs $E$, clean sample threshold $\tau$, hard sample threshold $\tau'$}

\tcp{Self-supervised pre-training} 
$f_{\phi}(\mathbf{x})$,$\{ \mathcal{N}_{\mathbf{x}_{i}} \}_{i=1}^{|\mathcal{D}|}$ = PreTrain($\mathcal{D}$) 

\tcp{Warm Up} 
$p_{\theta}(\mathbf{y}|\mathbf{x})$ = WarmUp($\mathcal{D}$,$f_{\phi}(\mathbf{x})$) \\
 \While{$e < E$}{
    \tcp{Estimate sets of clean and noisy samples}
    \For{$i = \{1,...,|\mathcal{D}|\}$}
 {Estimate $p(\text{clean}|\ell_i,\gamma)$, with $\ell_i=-\mathbf{y}_i^{\top}\log p_{\theta}(.|\mathbf{x}_i)$}
 
  $\mathcal{X},\mathcal{U}$=FormCleanNoisySets($\{p(\text{clean}|\ell_i,\gamma)\}_{i=1}^{|\mathcal{D}|},\tau$) 
  
 \tcp{Estimate sets of hard noisy and easy noisy samples}
    \For{$i = \{1,...,|\mathcal{U}|\}$}
 {Estimate $p \left ( \text{hard} |  \mathbf{y}_i^{*}(c),  \gamma \right )$, with $\mathbf{y}_i^{*}(c)=\argmax_{c \in \mathcal{Y}} p\left(c|\mathbf{x}_i \right)$}
 
 $\mathcal{U}_H, \mathcal{U}_E$=FormHardEasySets($\{p \left ( \text{hard} |  \mathbf{y}_i^{*}(c),  \gamma \right )\}_{i=1}^{|\mathcal{U}|},\tau'$) 
     \For{b=1 \textbf{to} $B$}   
            {
                % $\{(\mathbf{x}_b, \mathbf{y}_b, w_b)\}_{b=1}^{B} \sim (\mathcal{X}~U~\mathcal{U}_E)| w_b=0 ~\text{if}~ \mathbf{x}_b \in \mathcal{U}_E$\\
                
                $\{ \hat{\mathbf{x}}_{b,m}\}_{m=1}^{M}$ = DataAugment($\mathbf{x}_b \in \mathcal{X},M$)\\
                $\{ \hat{\mathbf{u}}_{b,m} \}_{m=1}^{M}$ = DataAugment($\mathbf{u}_b\in \mathcal{U}_E,M$)
                    
                $\mathbf{p}_b =  \frac{1}{2M}\sum\limits_{m=1,k=1}^{M,2}
                f(\hat{\mathbf{x}}_{b,m};\theta(k))$\\
                $\mathbf{q}_b = \frac{1}{2M}\sum\limits_{m=1,k=1}^{M,2} f(\hat{\mathbf{u}}_{b,m};\theta(k))$
                $\hat{\mathbf{y}}_b = \text{TempShrp}(w_b \mathbf{y}_b + (1-w_b)\mathbf{p}_b;T)$\\
                $\hat{\mathbf{q}}_b = \text{TempShrp}(\mathbf{q}_b;T)$
            }
$\mathcal{\hat{X}}=\{\hat{\mathbf{x}}_{b,m}, \hat{\mathbf{y}}_b\}_{m=1}^{M}$, 
$\mathcal{\hat{U}}_E=\{\hat{\mathbf{u}}_{b,m}, \hat{\mathbf{q}}_b\}_{m=1}^{M}$ \\
 $\mathcal{\hat{D}}$ = ProportionalMixUp($\mathcal{\hat{X}}, \mathcal{\hat{U}}_E$) \\
 Update $\theta(k1)$, $\theta(k2)$ with $\mathcal{L}$ from~(5)
 }
\caption{PropMix (PM) }
\label{alg:PM}
\end{algorithm}

% \bibliography{egbib}
\end{document}